\def\year{2021}\relax
\documentclass[letterpaper]{article} 
\usepackage{aaai21}  
\usepackage{times}  
\usepackage{helvet} 
\usepackage{courier}  
\usepackage[hyphens]{url}  
\usepackage{graphicx} 
\usepackage{natbib}  
\usepackage{caption} 

\usepackage{comment}
\usepackage{amsmath}
\usepackage{amssymb}

\usepackage{array}
\usepackage{booktabs}
\usepackage{multirow}
\usepackage[noend]{algpseudocode}
\usepackage[linesnumbered,ruled]{algorithm2e}

\usepackage[switch]{lineno}

\usepackage{hyperref}
\title{\textbf{Scalable NAS with Factorizable Architectural Parameters}}

\author{Lanfei Wang\textsuperscript{1}\thanks{This work was done when the first author was an intern at Huawei
Cloud \& AI BG.}\quad Lingxi Xie\textsuperscript{2}\quad Tianyi Zhang\textsuperscript{3}\quad Jun Guo\textsuperscript{1}\quad Qi Tian\textsuperscript{2}\\}

\affiliations{
\textsuperscript{1}Beijing University of Posts and Telecommunications\\
\textsuperscript{2}Huawei Cloud \& AI BG\quad
\textsuperscript{3}Tsinghua University\\
wanglanfei@bupt.edu.cn,
198808xc@gmail.com,
tianyi-z16@mails.tsinghua.edu.cn\\
guojun@bupt.edu.cn,
tian.qi1@huawei.com
}

\begin{document}

\maketitle
\begin{abstract}
Neural Architecture Search (NAS) is an emerging topic in machine learning and computer vision. The fundamental ideology of NAS is using an automatic mechanism to replace manual designs for exploring powerful network architectures. One of the key factors of NAS is to scale-up the search space, e.g., increasing the number of operators, so that more possibilities are covered, but existing search algorithms often get lost in a large number of operators. For avoiding huge computing and competition among similar operators in the same pool, this paper presents a scalable algorithm by factorizing a large set of candidate operators into smaller subspaces. As a practical example, this allows us to search for effective activation functions along with the regular operators including convolution, pooling, skip-connect, \textit{etc}. With a small increase in search costs and no extra costs in re-training, we find interesting architectures that were not explored before, and achieve state-of-the-art performance on CIFAR10 and ImageNet, two standard image classification benchmarks.
\end{abstract}

\section{Introduction}
\noindent Neural architecture search (NAS) is an important subarea of automated machine learning (AutoML). It aims to change the conventional, manual way of designing network architectures, and explore the possibility of finding effective architectures automatically in a large search space. Recent years have witnessed a blooming development of NAS, and automatically discovered architectures have surpassed manually designed ones in a wide range of vision tasks, \textit{e.g.}, image classification~\cite{zoph2018learning,liu2018progressive,TanL19}, object detection~\cite{GhiasiLL19,ElskenMH19}, semantic segmentation~\cite{LiuCSAHY019}, \textit{etc}. There are two main factors of NAS, namely, architecture sampling and evaluation. The heuristic search approaches, including using reinforcement learning~\cite{ZophL17,zoph2018learning,liu2018progressive} and genetic algorithms~\cite{RealMSSSTLK17,XieY17,real2019regularized}, often required an individual stage for evaluating each sampled architecture, while the so-called one-shot approaches~\cite{abs-1904-00420,liu2018darts,pham2018efficient} combined two stages into one so that the search process is much faster and thus can be applied to larger search spaces.

\begin{figure*}[!t]
\centering
\includegraphics[width=0.8\textwidth]{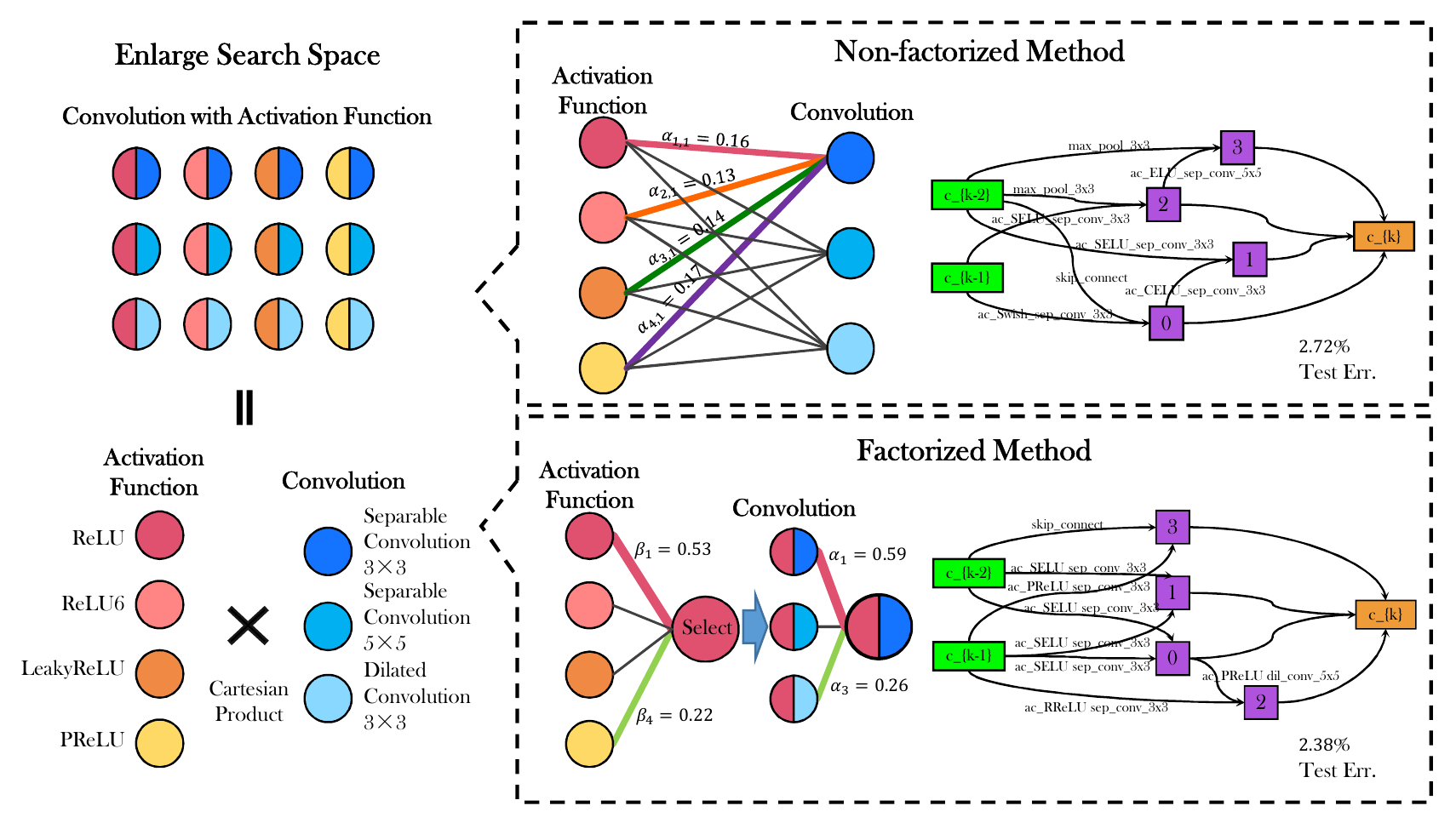} 
\caption{The ideology of this paper is to factorize a large search space (first row, top left, $\mathcal{O}_1\times\mathcal{O}_2$ operators) into two small spaces (second row, bottom left, $\mathcal{O}_1$ and $\mathcal{O}_2$ operators, respectively), so as to avoid the competition between similar combinations of operators (\textit{e.g.}, \textsf{ReLU \& sep-conv-3x3} vs. \textsf{PReLU \& sep-conv-3x3}). In each row, the middle column shows an example of weights learned by combination of operators (first row) or each operator (second row), and the right column shows the searched normal cell and its performance on CIFAR10.}
\label{fig:ideology}
\end{figure*}

This paper continues the efforts in scaling up the search space of NAS. We follow a recent algorithm named DARTS~\cite{liu2018darts}. By optimizing a super-network efficiently, DARTS has the potential of exploring a large space, \textit{e.g.}, in a cell with complex topology, each operator is an element in a regular set. However, as shown in Figure~\ref{fig:ideology}, if we hope to equip each operator with more attributes, \textit{e.g.}, the input of \textsf{sep-conv-3x3} can be pre-processed by either \textsf{ReLU} or \textsf{PReLU} activation, it can introduce more candidates into the search space, \textit{e.g.}, in the above example, both \textsf{ReLU \& sep-conv-3x3} and \textsf{PReLU \& sep-conv-3x3} are candidates. When these two `super-operators' compete in the same pool, the algorithm can be confused because their properties are very similar to each other. Moreover, when the number of such operators increases superlinearly, it often leads to heavy computational burdens.

Our idea of avoiding such competition is straightforward, \textit{i.e.}, \textbf{factorizing} the operators into two (or more) sets, so that each subspace does not contain very similar operators. In particular, we explore an example that the operator comes from the Cartesian product of an \textit{activation operator} group (\textit{e.g.}, \textsf{ReLU}, \textsf{PReLU}, \textsf{SELU}, \textit{etc.}) and a \textit{regular operator} group (\textit{e.g.}, \textsf{sep-conv-3x3}, \textsf{sep-conv-5x5}, \textsf{dil-conv-3x3}, \textit{etc.}). In other words, each pair of activation operator and regular operator can be combined, but the weights of all combinations are determined separately in the corresponding group. Consequently, the number of possibilities of the combined operators goes up quadratically as the number of trainable architectural parameters increasing linearly. It largely alleviates the computational burden as well as the instability of architecture search.

Our enlarged search space is defined by a factorizable set of operators, \textit{i.e.}, eight original regular operators (including \textsf{none}) and nine activation operators (including the originally default one, \textsf{ReLU}). Consequently, original DARTS~\cite{liu2018darts} space is enlarged from $1.1\times10^{18}$ to $9.3\times10^{29}$, enabling us to discover a lot of architectures that have not been studied before. Moreover, since the activation operators are efficient to calculate, the modified framework only needs $20\%$ extra overhead in the search stage, and no additional costs in both the re-training stage. Experiments are performed on CIFAR10 and ImageNet, on which our approach (A-DARTS) reports state-of-the-art accuracy ($97.62\%$ on CIFAR10, and $75.9\%$ top-1 on ImageNet), without either the SE-module or AutoAugment being integrated. When applied to PC-DARTS~\cite{xu2019pc}, a recently published variant of DARTS that directly searches on ImageNet, our approach (A-PC-DARTS) reports a higher top-1 accuracy of $76.2\%$, outperforming the baseline by a significant margin of $0.4\%$. Ablation experiments demonstrate that the improvement mostly comes from our `safe' search strategy in an enlarged space.

\section{Related Work}
\label{relatedwork}

Deep neural networks have been dominating in the field of computer vision. By stacking a large number of convolutional layers and carefully designing training strategies, manually designed networks~\cite{KrizhevskySH12,HeZRS16,huang2017densely} achieved great successes in a wide range of applications. However, the number of artificial architectures is still small, and it is widely believed that these architectures are prone to be sub-optimal. Recently, Neural Architecture Search (NAS) has attracted increasing attentions from both academia and industry~\cite{SandlerHZZC18,MobileNetV3}. As an emerging topic of AutoML, NAS originates from the motivation that powerful architectures can be found by automatic algorithms, provided that the search space is large enough~\cite{ZophL17}. Besides neural architectures, researchers also designed effective algorithms to search for other parameters, such as activation functions~\cite{RamachandranZL18} and data augmentation policies~\cite{CubukZMVL19}.

There are typically two pipelines of NAS, which differ from each other in how the search module interacts with the optimization module. The first pipeline isolated search from optimization, with the search module learning a parameterized policy of sampling architectures, and the optimization module providing rewards by training the sampled architecture from scratch. The search policy can be formulated with either reinforcement learning~\cite{ZophL17,zoph2018learning,pham2018efficient,CaiCZYW18} or evolutionary algorithms~\cite{RealMSSSTLK17,XieY17,real2019regularized}. Towards a higher search efficiency, researchers proposed to construct a super-network~\cite{zoph2018learning,tan2019mnasnet,WuDZWSWTVJK19,SandlerHZZC18,abs-1908-01748} to reduce the search space, or approximated the search space with modified strategies~\cite{MobileNetV3,liu2018progressive,ElskenMH18,ElskenMH19}.

However, the above pipeline suffers the difficulty of being computationally expensive, \textit{e.g.}, requiring hundreds or even thousands of GPU-days. To accelerate, researchers moved towards jointly optimizing the search and optimization modules. After efforts in sharing computation among individual optimization~\cite{CaiCZYW18,pham2018efficient}, the second pipeline, known as the one-shot method, was proposed. A representative work is DARTS~\cite{liu2018darts} which formulated architecture search in a differentiable manner so that end-to-end optimization was made possible. Despite efficiency, DARTS suffers the issue of instability, which was somewhat alleviated by follow-up variants~\cite{DongY19,chen2019progressive,abs-1908-01748,WuDZWSWTVJK19}, which mostly worked in DARTS search space. Besides, researchers worried that current differentiable search algorithms will get lost in a large space~\cite{abs-1910-11831,Understanding}. \textit{In this paper, we focus on investigating a relatively safe manner to enlarge the search space for the neural architecture search. This is the first work to factorize the space to scale up the differentiable search.}

\begin{figure*}[!t]
\centering
\includegraphics[width=0.8\textwidth]{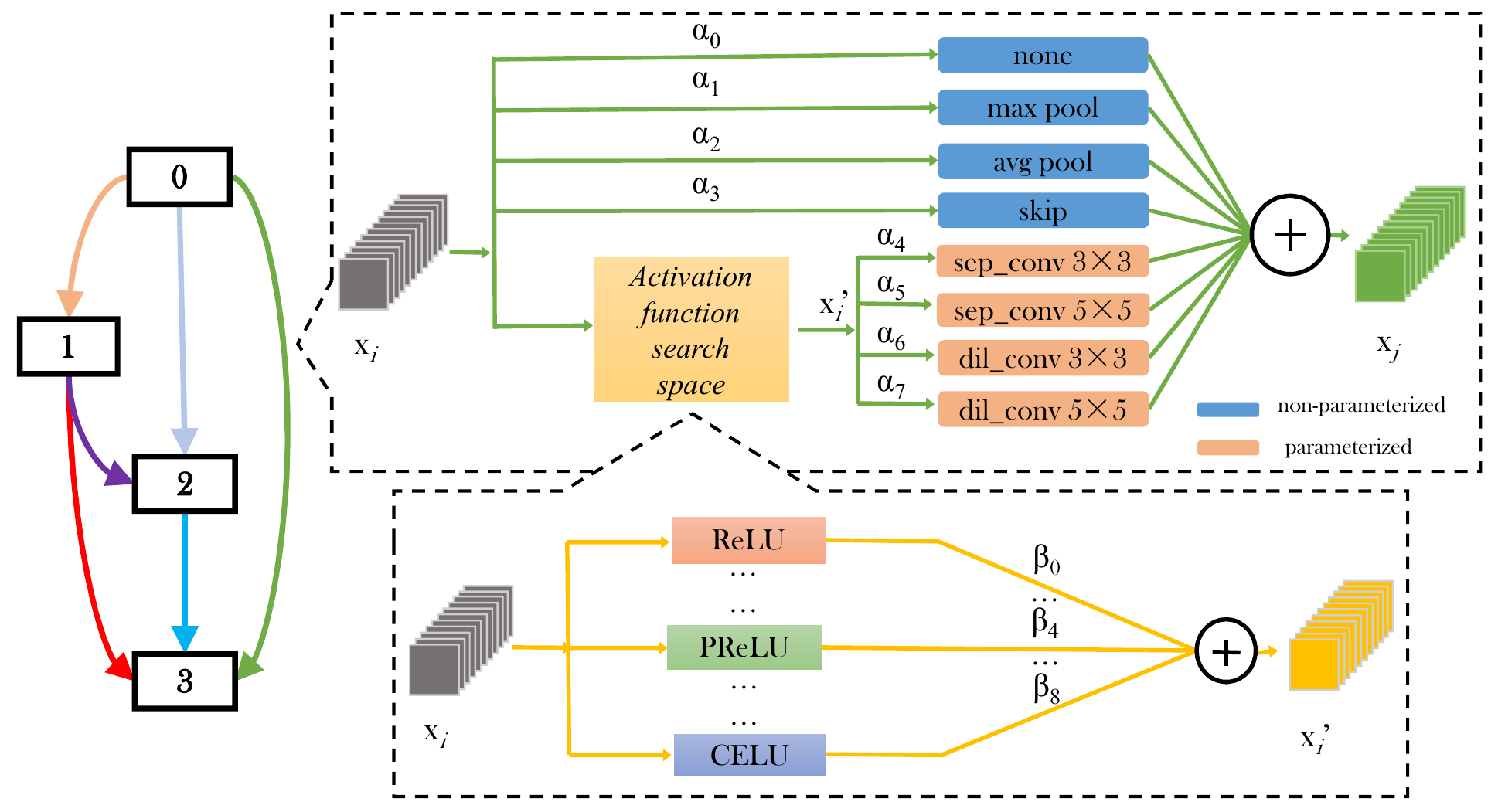} 
\caption{The factorized cell structure. On each edge of the cell, the \textit{regular operator} group is partitioned into two parts, \textit{i.e.}, parameterized and non-parameterized. The parameterized part is equipped with \textit{activation operator} group while the non-parameterized is not. The mixed-activation operator provides the input of the parameterized part.}
\label{fig:modified_cells}
\end{figure*}

\section{Our Approach}
\label{approach}

\subsection{Preliminaries: The DARTS Algorithm}

The fundamental idea of DARTS is to formulate NAS into a process of super-network optimization, in which the super-network is a mathematical function $f(\mathbf{x};\boldsymbol{\omega},\boldsymbol{\alpha})$, $\mathbf{x}$ denotes the input, $\boldsymbol{\omega}$ indicates the network parameters (\textit{e.g.}, convolutional weights) and $\boldsymbol{\alpha}$ indicates the architectural parameters (\textit{i.e.}, the weights of different operators). The goal is to find the optimal $\boldsymbol{\alpha}$ which determines the preserved sub-network (the final architecture), for which $\boldsymbol{\omega}$ is jointly trained to assist optimization.

The basic network of DARTS is stacked by a number of \textbf{cells}, each of which has a few nodes (indicating network layers with trainable operators) connected by the edges between them (indicating whether the output of one node is used as the input of another). In particular, let there be $N$ nodes in a cell, where $\left\{2,\ldots,{N-2}\right\}$ are intermediate nodes, and the default topology determines that node $i$ can send information to node $j$ if and only if $i<j$. Each intermediate node collects input data from the nodes with lower indices, and perform a mixed operation ${f_{i,j}\!\left(\mathbf{x}_i\right)}={{\sum_{o\in\mathcal{O}}}\mu_{i,j}^o\cdot o\!\left(\sigma\!\left(\mathbf{x}_i\right)\right)}$, where $\mathcal{O}$ is the set of candidate operators, $o$ is an element in it, and $\alpha_{i,j}^o$ is the architectural parameter on $o\!\left(\sigma\!\left(\mathbf{x}_i\right)\right)$ that determines the weight ${\mu_{i,j}^o}$, and ${\mu_{i,j}^o}={\exp\left\{\alpha_{i,j}^o\right\}/{\sum_{o'\in\mathcal{O}}}\exp\left\{\alpha_{i,j}^{o'}\right\}}$. We use $\sigma\!\left(\cdot\right)$ to denote the default activation function, \textit{i.e.}, \textsf{ReLU}~\cite{NairH10} (${\sigma\!\left(\mathbf{x}\right)}={\max\left\{\mathbf{x},\mathbf{0}\right\}}$, where $\max$ performs an element-wise operation). The output of a node is the sum of all input flows, \textit{i.e.}, ${\mathbf{x}_j}={{\sum_{i<j}}f_{i,j}\!\left(\mathbf{x}_i\right)}$, and the output of the last node $\mathbf{x}_{N-1}$ is the concatenation of the outputs of all intermediate nodes $\left\{\mathbf{x}_2,\ldots,\mathbf{x}_{N-2}\right\}$. Following the convention of DARTS, the first two nodes, $\mathbf{x}_0$ and $\mathbf{x}_1$, provide input and do not appear in the output of the cell.

\subsection{Enlarging the Search Space with a Factorizable Set of Regular and Activation Functions}

Our goal is to enlarge the space so that the search algorithm can explore more possibilities. To avoid confusion, we name the original group of operators (\textit{i.e.}, \textsf{sep-conv-3x3}, \textsf{sep-conv-5x5}, \textsf{dil-conv-3x3}, \textsf{sep-conv-5x5}, \textsf{max-pool-3x3}, \textsf{avg-pool-3x3}, \textsf{skip-connect}, and \textsf{none}) as \textit{regular operator} group,and try to allow the activation function
to change from the default one, \textsf{ReLU}, someone in the
activation operator group. The \textit{activation operator} group contains nine popular activation functions from the previous literature, namely, five variants of \textsf{ReLU} (\textit{i.e.}, \textsf{ReLU}, \textsf{ReLU6}~\cite{ReLU6}, \textsf{LeakyReLU}~\cite{LeakyReLU}, \textsf{PReLU}~\cite{HeZRS15} and \textsf{RReLU}~\cite{XuWCL15}, differing from the original version in the strategy of dealing with negative and saturating-positive responses), three variants of \textsf{ELU} (\textit{i.e.}, \textsf{ELU}~\cite{ClevertUH15}, \textsf{CELU}~\cite{Barron17a} and \textsf{SELU}~\cite{KlambauerUMH17}, also with subtle differences from each other), and \textsf{Swish}~\cite{RamachandranZL18}, which is in the form of $x\times\sigma\!\left(b\cdot x\right)$ where $\sigma\!\left(\cdot\right)$ is the sigmoid function and $b$ is the learned coefficient\footnote{Of course, it is possible to introduce other options such as the vanilla \textsf{sigmoid}, but we ignore them as they are often less effective.}.

This leads to a large search space. Since each activation function can be put in front of any convolutional operation, there are a total of ${9\times4+4}={40}$ `super-operators' in $\mathcal{O}$. Consequently, the search cost, which is proportional to $\left|\mathcal{O}\right|$, grows significantly. More importantly, DARTS enforces all operators to compete with each other (all operators share the total weight of $1.0$), but there exist candidates with very similar behaviors (\textit{e.g.}, \textsf{ReLU \& sep-conv-3x3} and \textsf{PReLU \& sep-conv-3x3}) that can confuse the search algorithm which must make a choice between them. In the experimental section, we will see that directly searching in such a space (referred to as non-factorized search) often leads to unsatisfying performance.

Our factorized method is intuitive and novel, \textit{i.e.}, decomposing the large operator set, $\mathcal{O}$, into the combination of two smaller sets, denoted as $\mathcal{O}_1$ and $\mathcal{O}_2$, so that ${\mathcal{O}}\subseteq{\mathcal{O}_1\times\mathcal{O}_2}$ where $\times$ denotes the Cartesian product. In the above example, $\mathcal{O}_1$ and $\mathcal{O}_2$ are the regular operator set and the activation operator set, respectively. This is named factorized search, which not only reduces the computational costs as well as the number of architectural parameters (${\left|\mathcal{O}_1\right|+\left|\mathcal{O}_2\right|}<{\left|\mathcal{O}_1\right|\times\left|\mathcal{O}_2\right|}$), but also alleviates the trouble that similar operators compete with each other.

\subsection{Tri-Level Optimization}

The factorized search space is illustrated in Figure~\ref{fig:modified_cells}. We use two individual sets of architecture parameters to formulate the super-network, $f\!\left(\mathbf{x};\boldsymbol{\omega},\boldsymbol{\alpha},\boldsymbol{\beta}\right)$, where $\boldsymbol{\alpha}$ and $\boldsymbol{\beta}$ denote the original and newly-added sets of architectural parameters, respectively. Hence, the overall mathematical function becomes:
\begin{equation}
\label{eqn:trilevel}
{f_{i,j}\!\left(\mathbf{x}_i\right)}={{\sum_{o_1\in\mathcal{O}_1}}\frac{\exp\left\{\alpha_{i,j}^{o_1}\right\}}{{\sum_{o_1'\in\mathcal{O}_1}}\exp\left\{\alpha_{i,j}^{o_1'}\right\}}\cdot o_1\!\left(\sigma_{i,j}^{o_1}\!\left(\mathbf{x}_i\right)\right)}.
\end{equation}
Note that the activation function has been replaced by $\sigma_{i,j}^{o_1}\!\left(\mathbf{x}_i\right)$, which is determined by both $o_1$ and $\boldsymbol{\beta}$. If $o_1$ falls within the non-parameterized subset of $\mathcal{O}_1$, no activation function is performed before $o_1\!\left(\cdot\right)$, we have ${\sigma_{i,j}^{o_1}\!\left(\mathbf{x}_i\right)}\equiv{\mathbf{x}_i}$, otherwise the output of activation follows a mixture formulation, which can be:
\begin{equation}
\label{eqn:trilevel2}
{\sigma_{i,j}^{o_1}\!\left(\mathbf{x}_i\right)}={{\sum_{o_2\in\mathcal{O}_2}}\frac{\exp\left\{\beta_{i,j}^{o_2}\right\}}{{\sum_{o_2'\in\mathcal{O}_2}}\exp\left\{\beta_{i,j}^{o_2'}\right\}}\cdot o_2\!\left(\mathbf{x}_i\right)}.
\end{equation}

On each edge $\left(i,j\right)$, the input feature, $\mathbf{x}_i$, gets through the parameters of $\boldsymbol{\beta}$, $\boldsymbol{\omega}$, and $\boldsymbol{\alpha}$ orderly and arrives at the output, $\mathbf{x}_j$. The overall loss function adopts a cross entropy loss function. Therefore, optimizing the super-network with respect to parameters ($\boldsymbol{\omega}$, $\boldsymbol{\alpha}$ and $\boldsymbol{\beta}$), which are updated by three optimizers in a gradient descent manner, involves a tri-level optimization process. We illustrate in Algorithm~\ref{alg:tri_level}.

\begin{algorithm}[!h]
\SetKwInOut{Input}{Input}
\SetKwInOut{Output}{Output}
\SetKwInOut{Return}{Return}
\Input{
a training set $\mathcal{T}$, a validation set $\mathcal{V}$, super-network $f\!\left(\cdot;\boldsymbol{\omega},\boldsymbol{\alpha},\boldsymbol{\beta}\right)$, loss functions $\mathcal{L}_\mathrm{train}\!\left(\cdot\right)$ and $\mathcal{L}_\mathrm{val}\!\left(\cdot\right)$, learning rates $\eta_{\boldsymbol{\alpha}}$, $\eta_{\boldsymbol{\beta}}$ and $\eta_{\boldsymbol{\omega}}$, number of search epochs $T$;
}
\Output{
parameters $\boldsymbol{\alpha}$ and $\boldsymbol{\beta}$;
}
Initialize $\boldsymbol{\alpha}_0$, $\boldsymbol{\beta}_0$ and $\boldsymbol{\omega}_0$, ${t}\leftarrow{0}$;\\
\While{\ ${t}<{T}$\ }{
${\boldsymbol{\omega}_{t+1}}\leftarrow{\boldsymbol{\omega}_t-\eta_{\boldsymbol{\omega}}\cdot\nabla_{\boldsymbol{\omega}_t}\mathcal{L}_\mathrm{train}\!\left(\boldsymbol{\omega}_t,\boldsymbol{\alpha}_t,\boldsymbol{\beta}_t\right)}$;\\
${\boldsymbol{\alpha}_{t+1}}\leftarrow{\boldsymbol{\alpha}_t-\eta_{\boldsymbol{\alpha}}\cdot\nabla_{\boldsymbol{\alpha}_t}\mathcal{L}_\mathrm{val}\!\left(\boldsymbol{\omega}_{t+1},\boldsymbol{\alpha}_t,\boldsymbol{\beta}_t\right)}$;\\
${\boldsymbol{\beta}_{t+1}}\leftarrow{\boldsymbol{\beta}_t-\eta_{\boldsymbol{\beta}}\cdot\nabla_{\boldsymbol{\beta}_t}\mathcal{L}_\mathrm{val}\!\left(\boldsymbol{\omega}_{t+1},\boldsymbol{\alpha}_{t+1},\boldsymbol{\beta}_t\right)}$;\\
${t}\leftarrow{t+1}$;\\
}
\Return{
${\boldsymbol{\alpha}}\leftarrow{\boldsymbol{\alpha}}_T$, ${\boldsymbol{\beta}}\leftarrow{\boldsymbol{\beta}}_T$.
}
\caption{Tri-Level Optimization for DARTS Equipped with Activation Function Search}
\label{alg:tri_level}
\end{algorithm}

\subsection{Discussion and Relationship to Previous Work}
The main contribution of our work is to provide an efficient way to expand the search space. Technically, the key is to avoid introducing a large number of architectural parameters. But, essentially, the idea is to assume the distribution of the hyper-parameters to be factorizable, so that we can use greedy search to optimize them individually. From this perspective, our idea is a little related to PNAS~\cite{liu2018progressive}, but there are big differences in search algorithm and optimization method. PNAS searched the space iteratively (\textit{i.e.}, optimizing one cell while freezing all others in each stage), however, we allow all operators to be jointly optimized.

\begin{table*}[!t]
\centering
\begin{tabular}{lccccc} 
\specialrule{0.05em}{0pt}{3pt} 
\multirow{2}{*}\textbf{Architecture}&\textbf{Test Err.}&\textbf{Params}&\textbf{Search Cost}&\textbf{Search GPU}&\multirow{2}{*}\textbf{Method}\\ 
\small &(\%)&(M) &(GPU-days) &(Type)\\
\specialrule{0.05em}{2pt}{2pt}
DenseNet-BC \cite{huang2017densely} &3.46 &25.6 &- &- &manual\\
\specialrule{0.05em}{2pt}{2pt}
NASNet-A~\cite{zoph2018learning}  &2.65 &3.3 &2000 &Tesla-P100 &RL\\
AmoebaNet-B~\cite{real2019regularized} &2.55$\pm$0.05 &2.8 &3150 &Tesla-K40 &evolution\\
ENAS~\cite{pham2018efficient}  &2.89 &4.6 &0.5 &GTX1080Ti &RL\\
\specialrule{0.05em}{2pt}{2pt}
DARTS (1st order)~\cite{liu2018darts}  &3.00$\pm$0.14 &3.3 &0.4 &GTX1080Ti &gradient\\
DARTS (2nd order)~\cite{liu2018darts}  &2.76$\pm$0.09 &3.3 &1 &GTX1080Ti &gradient\\
SNAS (moderate)~\cite{xie2018snas} &2.85$\pm$0.02 &2.8 &1.5 &Titan-XP &gradient\\
ProxylessNAS~\cite{cai2018proxylessnas} &2.08 &- &4.0 &- &gradient\\
P-DARTS~\cite{chen2019progressive}  &2.50 &3.4 &0.3 &Tesla-P100 &gradient\\
PC-DARTS~\cite{xu2019pc}  &2.57$\pm$0.07 &3.6 &0.06 &Tesla-V100 &gradient\\
\specialrule{0.05em}{2pt}{2pt}
A-DARTS (ours) &2.38$\pm$0.06 &3.7 &0.3 &Tesla-V100 &gradient\\
\specialrule{0.05em}{2pt}{0pt}
\end{tabular}
\caption{Comparison with state-of-the-art architectures on CIFAR10. The accuracy of our approach is reported by averaging $3$ individual runs of architecture search.}
\label{tab:CIFAR10}
\end{table*}

Furthermore, the activation functions search enjoys a natural benefit in the efficiency of the searched architectures. That being said, the time and memory costs do not change in the re-training stage. Yet, as we shall see later, the accuracy is significantly boosted. Previous work also explored this possibility and discovered \textsf{Swish}~\cite{RamachandranZL18}, but it was constrained that all activation functions are the same, whereas our work can assign different activation functions to different positions of the cell.

\section{Experiments}
\label{experiments}

\subsection{Results on CIFAR10}
Following DARTS~\cite{liu2018darts}, we use 8-cell super-network for searching on CIFAR10~\cite{article1} dataset, which contains $50\rm{,}000$ training and $10\rm{,}000$ testing images evenly distributed over $10$ classes. We set the basic channel number to be $16$, and use a batch size of $64$. The search epoch number is $25$ as the convergence of our approach. An SGD optimizer is used to optimize the weight parameters ($\boldsymbol{\omega}$), with an initial learning rate of $0.05$ and cosine annealed to a minimal value of $0.001$. The architectural parameters ($\boldsymbol{\alpha}$ and $\boldsymbol{\beta}$) are optimized by two Adam optimizers~\cite{KingmaB14} with the same configurations, \textit{i.e.}, a fixed learning rate of $6\times10^{-4}$, a momentum of $\left(0.5,0.999\right)$ and a weight decay of $10^{-3}$. 

After the search process, the final architecture is re-trained from scratch. We follow the convention~\cite{liu2018darts,chen2019progressive,xu2019pc} to use $18$ normal cells and $2$ reduction cells. The number of epochs is $600$, and the initial channel number is $36$. We set the batch size to be $128$, and use an SGD optimizer with an initial learning rate of $0.03$ (cosine annealed to $0$), a momentum of $0.9$, a weight decay of $3\times10^{-4}$ and a norm gradient client at $5$. {DropPath}~\cite{HuangSLSW16,LarssonMS17} and cutout~\cite{abs-1708-04552} are also used for re-training, with a dropout rate of $0.4$ and an auxiliary weight of $0.4$.

\noindent$\bullet$\quad\textbf{Comparison to the State-of-the-Arts}

The comparison of performance on CIFAR10 between our work and previous work is shown in Table~\ref{tab:CIFAR10}. With only a search cost of $0.3$ GPU-days, our approach dubbed A-DARTS achieves an error rate of $2.38\%$, which ranks among the top of existing DARTS-based approaches.

With factorization, our approach produces stable results on the enlarged space: in three individual search processes, the found architectures report $2.38\%$ (as shown in the table), $2.31\%$ and $2.46\%$ error rates, all of which are competitive. In addition, the searched architecture does not require longer time or larger memory in both re-training and testing stages. From the perspective of real-world applications, the improvement of accuracy comes for free.
 
\begin{figure*}[!t]
\centering
\begin{minipage}{0.38\linewidth}
\centerline	{\includegraphics[width=\textwidth]{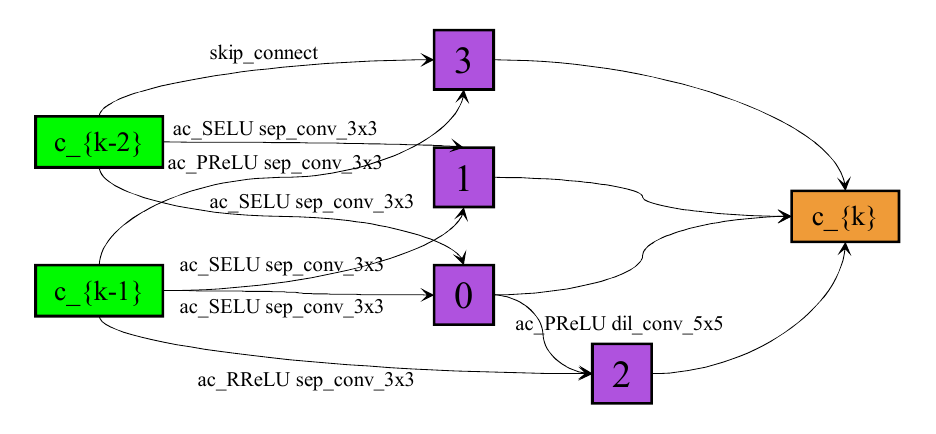}}
\centerline {\small(a) Normal cell on CIFAR10 (A-DARTS).}
\end{minipage}
\hspace{0.06\linewidth}
\begin{minipage}{0.38\linewidth}
\centerline	{\includegraphics[width=\textwidth]{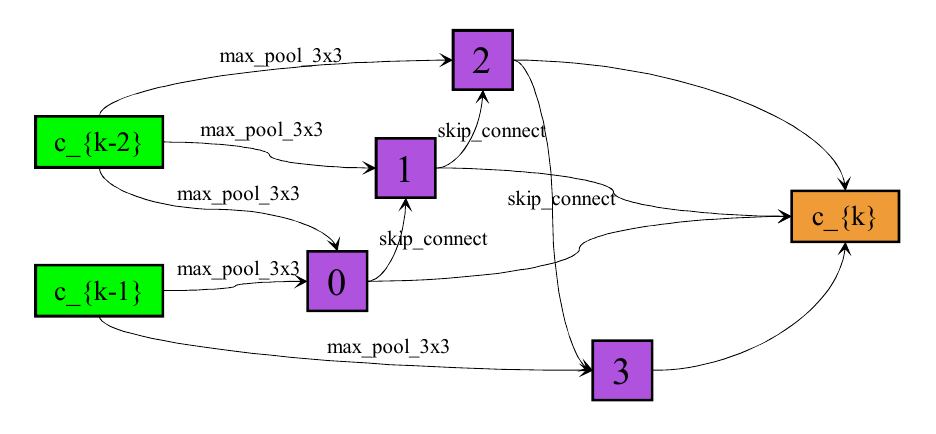}}
\centerline {\small(b) Reduction cell on CIFAR10 (A-DARTS).}
\end{minipage}
\vfill
\begin{minipage}{0.38\linewidth}
\centerline	{\includegraphics[width=\textwidth]{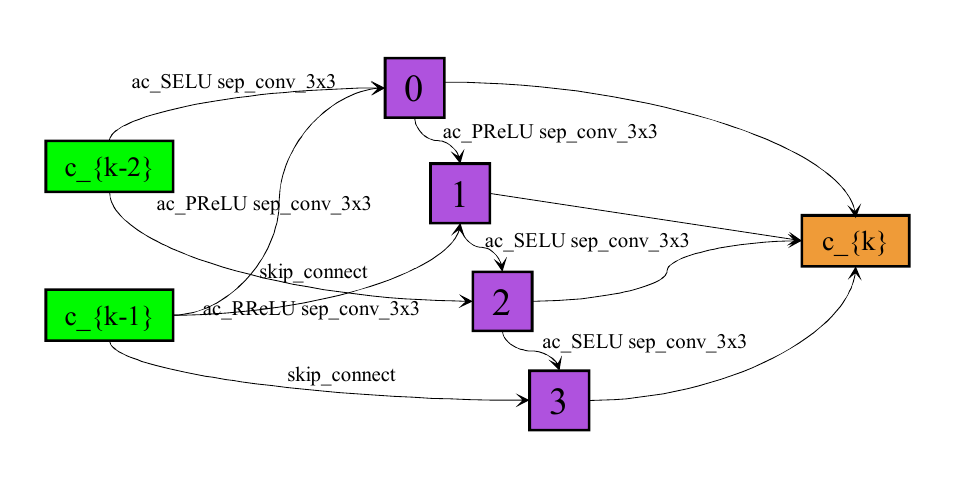}}
\centerline {\small(c) Normal cell on ImageNet (A-PC-DARTS).}
\end{minipage}
\hspace{0.06\linewidth}
\begin{minipage}{0.38\linewidth}
\centerline	{\includegraphics[width=\textwidth]{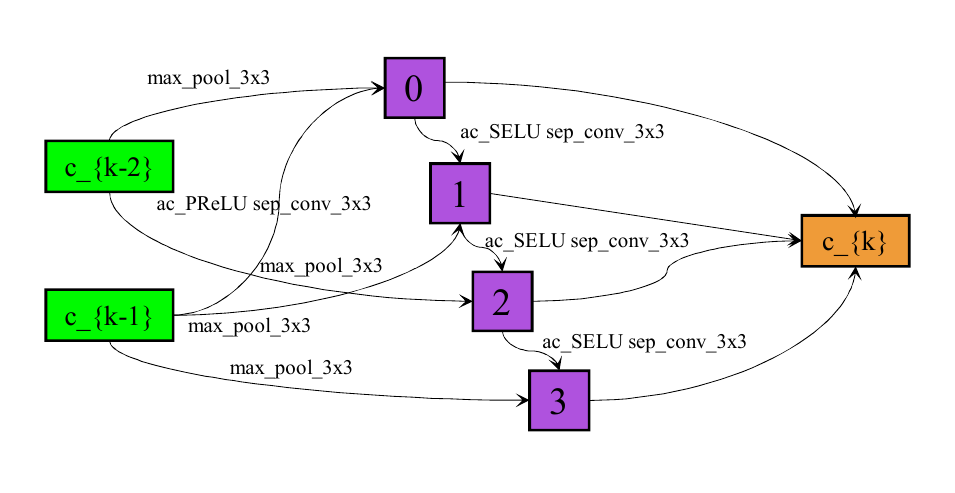}}
\centerline {\small(d) Reduction cell on ImageNet (A-PC-DARTS).}
\end{minipage}
\caption{The normal and reduction cells found by the factorized search method on CIFAR10 (\textbf{A-DARTS}, $0.3$ GPU-days, a test error of $2.38\%$) and ImageNet (\textbf{A-PC-DARTS}, $6.3$ GPU-days, a top-1 test error of $23.8\%$). More results are in the Appendix.}
\label{fig:visualization}       
\end{figure*}

The architecture corresponding to the error rate of $2.38\%$ is shown in the first row of Figure~\ref{fig:visualization} (a) and (b). In the searched normal cell, \textsf{SELU} appears in $4$ out of $7$ edges (one edge is occupied by \textsf{skip-connect} and thus does not need activation), claiming its dominance over other activation functions. In contrary, \textsf{Swish}~\cite{RamachandranZL18}, the searched activation function, does not appear in any edge, implying that \textsf{Swish} may not be the optimal choice in differentiable search (in particular, DARTS). As we shall see in the next part, even in the search using a fixed activation function, \textsf{Swish} does not show an advantage over other competitors.
 
\begin{table}[!t]
\centering
\setlength{\tabcolsep}{5pt}{
\begin{tabular}{cccc}
\specialrule{0.03em}{0pt}{3pt}
\multicolumn{2}{c}{\textbf{Configuration}}&\textbf{Test Err.}&\textbf{Params}\\
search with &re-train with &(\%) &(M)\\
\specialrule{0.03em}{2pt}{2pt}
\textit{factorized}      & \textit{factorized}      & \textbf{2.38}$\pm$0.06 & 3.7\\
\specialrule{0.03em}{2pt}{2pt}
\textsf{ReLU}      & \textsf{ReLU}      & 2.67$\pm$0.09 & 3.3\\
\textsf{ReLU6}     & \textsf{ReLU6}     & 2.65$\pm$0.08 & 3.8\\
\textsf{LeakyReLU} & \textsf{LeakyReLU} & 2.66$\pm$0.10 & 3.9\\
\textsf{PReLU}     & \textsf{PReLU}     & 2.62$\pm$0.10 & 3.6\\
\textsf{RReLU}     & \textsf{RReLU}     & 2.59$\pm$0.04 & 3.9\\
\textsf{ELU}       & \textsf{ELU}       & 2.62$\pm$0.08 & 3.6\\
\textsf{SELU}      & \textsf{SELU}      & 2.58$\pm$0.05 & 3.8\\
\textsf{CELU}      & \textsf{CELU}      & 2.62$\pm$0.05 & 4.0\\
\textsf{\textsf{Swish}}     & \textsf{\textsf{Swish}}     & 2.61$\pm$0.20 & 3.6\\
\specialrule{0.03em}{2pt}{2pt}
\textit{Config-A}  & \textit{Config-A}  & 2.50$\pm$0.14 & 4.0\\
\specialrule{0.03em}{2pt}{2pt}
\textit{factorized}      & \textsf{ReLU}      & 2.50$\pm$0.05 & 3.7\\
\textit{factorized}      & \textsf{SELU}      & 2.42$\pm$0.06 & 3.7\\
\specialrule{0.03em}{2pt}{2pt}
\textit{non-factorized}      & \textit{non-factorized}      & 2.72$\pm$0.10 & 3.4\\
\specialrule{0.03em}{2pt}{2pt}
\end{tabular}}
\caption{Comparison of performance on CIFAR10, between our approach and those with different options, \textit{e.g.}, in the 2nd and 3rd rows from the bottom, we retrain the model by replacing the searched activation functions with the fixed ones. The search cost is is $0.25$ GPU-days for nine fixed activation and config-A, and non-factorized one requires $4.2$ GPU-days, others need $0.3$ GPU-days.}
\label{tab:fixed_activation_CIFAR10}
\end{table}
 
\noindent
$\bullet$\quad\textbf{Comparisons to Other Search Strategies}

In this part, we perform diagnostic studies to verify the effectiveness of searching in the enlarged space. \textbf{First}, we enumerate all $9$ elements in the activation function set and re-run the search process with each of them fixed to be the default activation. As shown in Table~\ref{tab:fixed_activation_CIFAR10}, all these activation functions produce statistically comparable results, with the worst being \textsf{ReLU}, and the best being \textsf{SELU}, \textit{i.e.}, the one that appears most frequently in the searched architecture. However, searching with \textsf{SELU} as the fixed activation reports an error rate of $2.58\%$, $0.2\%$ higher than our strategy, implying the benefit of adopting our factorized method in architecture search with activation function search.

\textbf{Second}, we leverage the configuration of activation functions from the best architecture (a $2.38\%$ error, which we call \textit{Config-A}), fix the architectural parameters of activation functions (the $\boldsymbol{\beta}$ at the end of the previous search) and redo the search with $\boldsymbol{\alpha}$ and $\boldsymbol{\omega}$ learnable. This reports a slightly higher error of $2.50\%$. On the other hand, if we directly use the \textit{factorized} architecture but simply replace the corresponding activations to fixed ones in the re-training stage, all \textsf{ReLU} and all \textsf{SELU}, respectively, the accuracy drops accordingly, and the drop caused by \textsf{SELU} is much smaller (note that \textsf{SELU} is stronger than \textsf{ReLU} in fixed-activation settings). These experiments indicate the benefit of searching in an enlarged space. During the search process, the two sets of architectural parameters impact each other, and the optimal architecture is not likely to be found if we fix either part and search only on the reduced space. That being said, being able to explore in a large space is more important than the results obtained from the large space. We acquire this ability by factorizing architectural parameters, which is currently a powerful yet safe solution.

\textbf{Third} and most importantly, we evaluate the option of non-factorized search, in which all possible combinations of activation functions and regular operators compete in the same pool. This strategy leads to a large number of responses to be computed and thus the search cost becomes high, $4.2$ GPU-days on CIFAR10, yet the searched architecture reports an error rate of $2.72\%$. In comparison, factorized search reports a $2.38\%$ error in $0.3$ GPU-days. This verifies that factorized search is efficient yet avoids performance deterioration caused by the competition among similar candidate operators. Please refer to the Appendix for further analysis and visualization of the non-factorized search.

\begin{table*}[!t]
\centering
\setlength{\tabcolsep}{0.16cm}
\begin{tabular}{lcccccc} 
\specialrule{0.05em}{0pt}{3pt} 
\multirow{2}{*}\textbf{Architecture}&\multicolumn{2}{c}{\textbf{Test Err.(\%)}}&\textbf{Params}&\textbf{FLOPs}&\textbf{Search Cost}&\multirow{2}{*}\textbf{Method}\\ 
&top-1&top-5&(M) &(M) &(GPU-days)&\\
\specialrule{0.05em}{2pt}{2pt}
Inception-v1 \cite{szegedy2015going} &30.2 &10.1 &6.6 &1448 &- &manual\\
MobileNet \cite{howard2017mobilenets} &29.4 &10.5 &4.2 &569 &- &manual\\
ShuffleNet 2 $\times$ (v2) \cite{ma2018shufflenet} &25.1 &- &$\sim$5 &591 &- &manual\\
\specialrule{0.05em}{2pt}{2pt}
NASNet-A \cite{zoph2018learning} &26.0 &8.4 &5.3 &564 &1800 &RL\\
AmoebaNet-C \cite{real2019regularized} &24.3 &7.6 &6.4 &570 &3150 &evolution\\
PNAS \cite{liu2018progressive} &25.8 &8.1 &5.1 &588 &225 &SMBO\\
MnasNet-92 \cite{tan2019mnasnet} &25.2 &8.0 &4.4 &388 &- &RL\\
\specialrule{0.05em}{2pt}{2pt}
SNAS (mild) \cite{xie2018snas} &27.3 &9.2 &4.3 &522 &1.5 &gradient\\
ProxylessNAS\textsuperscript{\ddag} \cite{cai2018proxylessnas} &24.9 &7.5 &7.1 &465 &8.3 &gradient\\
DARTS (2nd order) \cite{liu2018darts} &26.7 &8.7 &4.7 &574 &4.0 &gradient\\
P-DARTS (CIFAR10) \cite{chen2019progressive} &24.4 &7.4 &4.9 &557 &0.3 &gradient\\
PC-DARTS\textsuperscript{\ddag} \cite{xu2019pc} &24.2 &7.3 &5.3 &597 &3.8 &gradient\\
\specialrule{0.05em}{2pt}{2pt}
A-DARTS (ours) &24.1 &7.3 &5.2 &587 &0.3 &gradient\\
A-PC-DARTS\textsuperscript{\ddag} (ours) &23.8 &7.0 &5.3 &598 &6.3 &gradient\\
\specialrule{0.05em}{2pt}{0pt}
\end{tabular}
\caption{Comparison with state-of-the-art network architectures on ImageNet (mobile setting). The symbol of \textsuperscript{\ddag} indicates the corresponding architecture is searched on ImageNet directly. Otherwise it is transferred from CIFAR10.}
\label{tab:ImageNet}
\end{table*}

\subsection{Results on ImageNet}
We adopt two ways to obtain architectures for ImageNet~\cite{DengDSLL009}, exactly, ILSVRC2012~\cite{RussakovskyDSKS15}, which has $1\rm{,}000$ object classes and around $1.3\mathrm{M}$ training and $50\mathrm{K}$ testing images. One is to search the architecture on a \textit{proxy} dataset then transfer it to ImageNet, and the other is to directly search the architecture on ImageNet, known as the \textit{proxyless} setting~\cite{cai2018proxylessnas,xu2019pc}.

Regarding the proxyless setting (\textit{i.e.}, search on ImageNet directly), we follow PC-DARTS~\cite{xu2019pc} to sample a subset from training set, with the training and validation set containing $10\%$ and $2.5\%$ images, respectively, and all classes preserved. A total of $50$ epochs are used for searching, and the first $15$ epochs are used for warm-up. To train the network weights $\boldsymbol{\omega}$, a momentum SGD is used, with the learning rate starting from $0.5$ and gradually annealed down to $0$ in a cosine schedule. The momentum is $0.9$, the weight decay is $3\times10^{-5}$. To train the architectural weights, $\boldsymbol{\alpha}$ and $\boldsymbol{\beta}$, we use two Adam optimizers~\cite{KingmaB14} with a fixed learning rate of $6\times10^{-3}$, a momentum $(0.5,0.999)$ and a weight decay of $10^{-3}$. We search on $8$ NVIDIA Tesla-V100 GPUs. Following PC-DARTS to sub-sample $1/2$ channels at each time, the batch size can be as large as $1\rm{,}024$. The entire search process takes $19$ hours, slightly longer than PC-DARTS.

For both proxy and proxyless settings, the searched architectures have $8$ cells. We deepen the network by doubling the number of normal cells at each stage, resulting in a $14$-cell network.The number of initial channels is computed so that the total FLOPs does not exceed 600M.

We use $8$ NVIDIA Tesla-V100 GPUs to re-train the transferred architectures for $250$ epochs from scratch, with the batch size of $512$. The network parameters are optimized by a SGD optimizer with an initial learning rate of $0.25$ (gradually annealed to close to $0$ in cosine way), a momentum of $0.9$ and a weight decay of $3\times10^{-5}$. Default enhancements are also used, including label smoothing~\cite{SzegedyVISW16} and an auxiliary loss tower with $0.2$ auxiliary weight. The learning rate rises linearly to $0.25$ in the first $5$ epochs, which called a warm-up stage.

\noindent$\bullet$\quad\textbf{Comparison to the State-of-the-Arts}

Results are summarized in Table~\ref{tab:ImageNet}. In both the proxy and proxyless settings, our approach demonstrates state-of-the-art performance. In particular, in the proxy setting, the top-1 error rate of $24.1\%$ surpasses all published DARTS-based results. Note that A-DARTS was built on the original DARTS algorithm, when we combine our approach with PC-DARTS, which was published recently, the classification error of A-PC-DARTS continues to drop. Without bells and whistles (e.g., the squeeze-and-excitation~\cite{HuSS18} module and AutoAugment~\cite{CubukZMVL19} which were verified effective), we arrive at a top-1 error of $23.8\%$, which surpasses PC-DARTS by a margin of $0.4\%$. This verifies that the idea of factorizing the search space is generalized to other search algorithms.

The searched architectures, using A-PC-DARTS, are visualized in Figure~\ref{fig:visualization}. Unlike those found on CIFAR10, this architecture has much deeper connections, \textit{i.e.}, $4$ cascaded convolutional layers exist in both normal cell and reduction cell. Also, the appearance of activation functions is similar to that observed in CIFAR10, \textit{e.g.}, \textsf{SELU} is chosen most frequently, both \textsf{PReLU} and \textsf{RReLU} contribute to the improvement, and \textsf{ReLU} is not chosen by any edge. This implies that some activation functions are prone to be selected by differentiable search, DARTS-based algorithms, exactly.

\begin{table}[!h]
\centering
\setlength{\tabcolsep}{0.3pt}{
\begin{tabular}{cccccc}
\specialrule{0.05em}{0pt}{3pt}
\multicolumn{2}{c}{\textbf{Configuration}}&\multicolumn{2}{c}{\textbf{Test Err. (\%)}}&\textbf{Params}&\textbf{FLOPs}\\
search with &re-train with &top-1&top-5 &(M) &(M)\\
\specialrule{0.05em}{2pt}{2pt}
\textit{factorized}      & \textit{factorized}   &\textbf{24.1} &7.3 &5.2 &587\\
\specialrule{0.05em}{2pt}{2pt}
\textsf{ReLU}      & \textsf{ReLU}  &24.9 &7.4 &5.2 &583\\
\textsf{SELU}      & \textsf{SELU}  &24.6 &7.3 &5.1 &584\\
\specialrule{0.05em}{2pt}{2pt}
\textit{factorized}      & \textsf{ReLU}   &24.5 &7.4 &5.2 &587\\   
\textit{factorized}      & \textsf{SELU}   &24.3 &7.3 &5.2 &587\\
\specialrule{0.05em}{2pt}{2pt}
\textit{non-factorized}      & \textit{non-factorized}   &24.6  &7.5  &5.2 &589\\
\specialrule{0.05em}{2pt}{2pt}
\end{tabular}}
\caption{Comparison of performance on ImageNet, between our approach and those with different options. Please refer to Table~\ref{tab:fixed_activation_CIFAR10} for the corresponding results on CIFAR10.}
\label{tab:fixed_activation_ImageNet}
\end{table}

\noindent
$\bullet$\quad\textbf{Comparisons to Other Search Strategies}

We perform experiments to evaluate other search strategies, and summarize results in Table~\ref{tab:fixed_activation_ImageNet}. Again, the factorized search strategy outperforms other strategies, including using a fixed activation function to search and retrain, and replacing the optimal activations with fixed ones to retrain. Also, searching in the non-factorizable space reports worse results but requires heavier computational costs. These results align with the conclusions made in CIFAR10 experiments.

\section{Conclusions}
\label{conclusions}

In this paper, we propose an efficient yet effective approach to enlarge the search space. Our factorized method makes the number of combinations increases super-linearly with the learnable architectural parameters increasing linearly, which reduces both computational overhead and optimization difficulty. Experiments on CIFAR10 and ImageNet demonstrate superior performance beyond the state-of-the-art of DARTS-based approaches, and diagnostic experiments verify the benefits of the factorized search method.

Our research provides a new way to expand search space without incurring instability~\cite{abs-1910-11831,Understanding}. Our work leaves many issues to be solved. For example, convolutional operations have orthogonal properties, \textit{e.g.}, different convolution types (\textit{e.g.}, normal, separable, dilated, \textit{etc.}) combined with different kernel sizes, and factorizing them into smaller sets can improve automation of architecture search, which meets the intention of AutoML~\cite{automl}. Topics are left for future work.

\section*{Ethical Impact}
This paper is our own original work, which presents a scalable NAS algorithm and allows activation functions to be searched with regular operators, We summarize the potential impact of our work in the following aspects.
\begin{itemize}
\item \textbf{To the research community.} We provide a safe and efficient solution for enlarging the search space. This is important for the NAS community, because we believe that exploring a large space is a fundamental ability of NAS algorithms. As a preliminary example, we allow activation functions to be searched together with regular operations, which achieves consistently better accuracy than in the original space.
\item \textbf{To the downstream engineers.} Our algorithm is fast and produces higher accuracy on standard classification datasets. After the code is released, there may be some engineers that deploy our algorithm to their own tasks. While this may help them in developing AI-based applications, there exist risks that some engineers, with relatively less knowledge in deep learning, can deliberately use the algorithm, \textit{e.g.}, introducing some improper operators to the search space, which may actually harm the performance of the designed system.
\item \textbf{To society.} There is a long-lasting debate on the impact that AI can bring to human society. Being an algorithm for improving the fundamental ability of deep learning, our work lies on the path of advancing AI. Therefore, in general, it can bring both beneficial and harmful impacts and it really depends on the motivation of the users.
\end{itemize}

We also encourage the community to investigate the following problems.
\begin{enumerate}
\item Is it possible to integrate other kinds of operators (\textit{e.g.}, variants of batch normalization)? Can these add-ons be jointly searched?
\item Is it possible to factorize the set of regular operators (\textit{e.g.}, for convolution, according to the kernel size and the type of convolution) to support a larger number of regular operators?
\item Is the choice among different activation functions identical when the algorithm is applied to other kinds of vision tasks? If not, what will be the difference?
\end{enumerate}

\bibstyle{aaai}
\bibliography{egbib}

\end{document}